\documentclass[11pt,a4paper]{article}
\usepackage[utf8]{inputenc}
\usepackage[T1]{fontenc}
\usepackage{amsmath,amssymb,amsthm}
\usepackage{graphicx}
\usepackage{booktabs}
\usepackage{hyperref}
\usepackage{geometry}
\usepackage{times}
\usepackage{caption}
\usepackage{mathtools}
\usepackage{url}
\usepackage{comment}

\usepackage{lastpage}

\begin{document}

\title{Breaking Quadratic Barriers: A Non-Attention LLM for Ultra-Long Context Horizons}

\author{Andrew Kiruluta, Preethi Raju  and Priscilla Burity \\ \small School of Information \\ \small UC Berkeley, CA}

\maketitle

\begin{abstract}
We present a novel non-attention-based architecture for large language models (LLMs) that efficiently handles very long context windows, on the order of hundreds of thousands to potentially  millions of  tokens. Unlike traditional Transformer designs, which suffer from $\mathcal{O}(n^2)$ memory and computation overload due to the nature of the self-attention mechanism, our model avoids token-to-token attention entirely. Instead, it combines the following complementary components: State-Space blocks (inspired by S4) that learn continuous-time convolution kernels and scale near-linearly with sequence length, Multi-Resolution Convolution layers that capture local context at different dilation levels, a lightweight Recurrent Supervisor to maintain a global hidden state across sequential chunks, and Retrieval-Augmented External Memory that stores and retrieves high-level chunk embeddings without reintroducing quadratic operations. Extensive analysis shows that this design naturally sidesteps the $\mathbf{QK}^\top\mathbf{V}$ attention matrix, resulting in significant computational and memory savings for ultra-long sequences. Benchmarks on WikiText-103 and Enwik8 demonstrate competitive or superior performance compared to efficient transformer variants and other non-attention approaches, while retaining the ability to scale beyond the limits of conventional attention. As a result, our model offers a compelling alternative for tasks requiring massive context windows, such as multi-document summarization, long-horizon question answering, and codebase understanding.

\end{abstract}


\section{Introduction: The Evolving Landscape of Language Modeling}

\subsection{The Rise of Transformer Networks and the Challenge of Context Length}
The advent of Large Language Models (LLMs) has marked a paradigm shift in natural language processing, unlocking unprecedented capabilities in tasks ranging from coherent text generation and document summarization to code synthesis and sophisticated in-context learning (\cite{RadfordOpenAI}; \cite{DevlinBERT}; \cite{BrownFewShot}; \cite{OpenAIGPT4}). At the heart of this revolution lies the Transformer architecture (\cite{VaswaniAttention}), with its core mechanism of self-attention enabling the modeling of intricate dependencies between tokens within a sequence. The remarkable success of models like the GPT series (Radford et al., 2019) and T5 (Raffel et al., 2020) underscores the power of the $\mathbf{QK}^\top\mathbf{V}$ attention operation in capturing contextual relationships.

However, this very operation, while crucial for expressivity, introduces a significant computational and memory bottleneck. The self-attention mechanism scales quadratically with the sequence length $n$, rendering it increasingly challenging to process or reason over very long contexts. While early LLMs were limited to context windows of a few thousand tokens (e.g., GPT-3's 2k tokens [Brown et al., 2020]), the demand for models capable of handling context windows spanning hundreds of thousands, or even millions, of tokens has grown exponentially. This need arises from various real-world applications where the relevant information is distributed across vast amounts of text, such as analyzing legal documents, scientific literature, or extensive software repositories.

\subsection{The Quest for Extended Context: Limitations of Attention-Based Approaches}
In response to the context length challenge, researchers have explored various strategies to extend the capabilities of Transformer networks. One prominent direction involves the development of efficient attention mechanisms. Approaches like BigBird (\cite{ZaheerBigBird}) and Longformer (\cite{BeltagyLongformer}) constrain the attention of each token to a limited set of other tokens, employing local windows or sparse patterns. These techniques achieve a theoretical complexity of $\mathcal{O}(n)$ or $\mathcal{O}(n \log n)$ in the best case, while striving to retain the performance benefits of full self-attention. Nevertheless, when confronted with truly massive contexts, on the order of hundreds of thousands or millions of tokens, these sparse or block-wise attention patterns can still lead to substantial computational and memory overhead, particularly when implementing complex local/global attention schemes. Furthermore, the effectiveness of these sparse patterns can be task-dependent and may require careful tuning.

Another avenue of research focuses on approximating the self-attention mechanism. Kernel-based methods, such as Performer (\cite{ChoromanskiPerformer}), utilize random Fourier features or other kernel expansions to linearize the attention operation, achieving a complexity of $\mathcal{O}(n)$. Similarly, low-rank approaches like Linformer (Wang et al., 2020) project the sequence dimension into a lower-dimensional subspace, thereby reducing the size of the attention map. While these approximation techniques can significantly reduce computational costs, their performance can be sensitive to the choice of approximation rank or kernel representation, and they fundamentally remain rooted in the concept of attention.

Recurrent or segment-based Transformer variants, such as Transformer-XL (\cite{DaiTXL}, 2019) and the Compressive Transformer (\cite{RaeCompress}), offer another approach. These models process the input sequence in segments and cache hidden states from previous segments, enabling the model to access information from a potentially unlimited context without recomputing full attention at each step. However, managing and compressing these cached states over millions of tokens presents its own set of challenges and can involve intricate mechanisms for deciding which information to retain or discard.

\section{Related Work}

\subsection{The DeepSeek LLM: A Contemporary Effort in Context Extension}
A recent development in the pursuit of extended context windows is the DeepSeek LLM (\cite{GuoDeepSeek}), reportedly developed by a Chinese research group. This model aims to push the boundaries of context length beyond the thousands of tokens by employing a multi-stage chunk processing approach combined with advanced caching and memory mechanisms. While the precise architectural details of DeepSeek LLM are still emerging, early discussions suggest that it relies on an extended Transformer backbone or a "hybrid" approach that likely incorporates some form of attention-based mechanism, potentially at specific layers or across chunk boundaries, to facilitate information flow across large contexts. While such design choices may indeed enable the processing of longer sequences compared to classical Transformers, DeepSeek LLM appears to retain some reliance on the fundamental principles of self-attention for integrating information across different segments of the input.

\subsection{A Paradigm Shift: Our Attention-Free Approach}
The approach we propose in this paper represents a significant departure from the strategies employed by DeepSeek LLM and other contemporary efforts focused on extending the context window of language models. Our architecture fundamentally eliminates the self-attention mechanism from its design. Instead, we introduce a synergistic combination of State-Space Model (SSM) inspired blocks, multi-resolution convolutional layers, a global recurrent supervisor, and a retrieval-augmented external memory. In essence, our model does not rely on the $\mathbf{QK}^\top\mathbf{V}$ computation at any stage of its processing. This fundamental shift in architecture results in a computational complexity that scales near-linearly or as $\mathcal{O}(n \log n)$ with the sequence length, directly addressing the ultra-long context challenge tackled by DeepSeek LLM, but through an entirely different and novel mechanism.

While the specific internal workings of DeepSeek LLM are still being elucidated, it appears to maintain or approximate the self-attention paradigm to some extent. In contrast, our proposed architecture discards attention entirely and organizes its processing pipeline around chunked state-space transformations and retrieval-based bridging of contextual information. This fundamental difference underscores the novelty of our approach, even in the context of recent advancements like DeepSeek LLM. By completely eschewing attention, we directly circumvent the quadratic scaling issues and the need for complex approximation methods. We believe that our approach is complementary to any method that still utilizes partial attention. Should DeepSeek LLM's method eventually encounter bottlenecks related to large-scale attention across segments, our purely attention-free pipeline might offer a more scalable and efficient blueprint for future ultra-long context language models.

\subsection{The Rationale for an Attention-Free LLM}
The self-attention mechanism empowers Transformer networks with remarkable expressivity, enabling the model to focus on arbitrary pairs of tokens within a sequence, regardless of their distance. However, this flexibility comes at a significant cost:

\begin{itemize}
    \item \textbf{$\mathcal{O}(n^2)$ Scalability Challenge:} The quadratic scaling of self-attention with sequence length $n$ renders it computationally infeasible for million-token contexts.
    \item Memory Footprint: Storing and manipulating the large $n \times n$ attention maps quickly exhausts the memory capacity of even the most powerful GPUs and TPUs.
    \item Trade-Offs: Techniques like sparse attention or kernel-based approximations, while reducing computational cost, can potentially degrade the accuracy of the model or still incur significant overhead at extremely large values of $n$.
\end{itemize}

In contrast, a non-attentional architecture, leveraging state-space operators, convolution, and a recurrent global memory, naturally aligns with sub-quadratic complexity. Our proposed chunked processing approach, coupled with a retrieval-augmented external memory, exhibits the potential to scale fluidly to millions of tokens, enabling tasks that involve processing entire corpora or massive codebases. Furthermore, by eliminating the engineering complexities associated with specialized attention patterns, our approach focuses on simpler and potentially more efficient convolution-based transformations and explicit memory retrieval mechanisms.

\subsection{Key Motivations Driving Our Research}
Our pursuit of an attention-free LLM for ultra-long contexts is driven by several key motivations:

\begin{itemize}
    \item Processing Massive Documents: Many real-world documents, such as legal texts, regulatory filings, and scientific articles, often exceed 100,000 tokens each. The ability to summarize, analyze, or reason over these documents as unified sequences has the potential to be transformative but remains largely unattainable with standard attention-limited LLMs.
    \item Understanding Large Codebases: Software repositories containing hundreds of thousands of lines of code necessitate the ability to understand cross-file dependencies and maintain a global context. A purely chunk-based approach can struggle to track distant interactions unless the context window is exceptionally large or a robust bridging mechanism is in place. Our proposed retrieval mechanism, in conjunction with the synergistic properties of state-space models, directly addresses this challenge.
    \item Enabling Multi-Document Question Answering: The capability to merge multiple documents and query them in a single forward pass can significantly reduce the reliance on complex pipeline approaches that involve chopping data and performing repeated partial inferences.
    \item Reducing Computational Costs: By discarding the computationally intensive $\mathbf{QK}^\top\mathbf{V}$ operation, we drastically reduce the memory usage associated with processing long sequences, opening up new avenues for research and development without requiring access to massive high-performance computing (HPC) clusters.
\end{itemize}

Therefore, while methods like DeepSeek LLM and existing efficient Transformer variants demonstrate a growing recognition of the need for ultra-long context processing, our proposed attention-free pipeline distinguishes itself through its complete elimination of self-attention and its focus on achieving near-linear scaling. This unique design offers a more direct and potentially more scalable solution to the inherent computational and memory challenges associated with extended context windows, while maintaining state-of-the-art performance on tasks that have historically relied on robust token-to-token interactions.

\subsection{Approximation and Caching Based Techniques: A Closer Look}
Beyond sparse attention, another significant area of research aimed at mitigating the $\mathcal{O}(n^2)$ cost of self-attention involves approximating the softmax function or caching partial attention states. Kernel-based approaches, exemplified by Performer (\cite{ChoromanskiPerformer}), employ random Fourier features or other kernel expansions to achieve a linear complexity for the attention mechanism, resulting in $\mathcal{O}(n)$ or $\mathcal{O}(n \log n)$ time complexity. Concurrently, methods like Linformer (Wang et al., 2020) project the sequence dimension into a lower-dimensional subspace, effectively reducing the size of the attention map and thus the computational cost. While these techniques offer substantial reductions in computational overhead, their performance can be compromised if the approximation rank or kernel representation is insufficient, and they still fundamentally rely on the concept of attention as their core mechanism.

In parallel, recurrent or segment-based Transformer variants, such as Transformer-XL (\cite{DaiTXL}, 2019) and the Compressive Transformer (\cite{RaeCompress}), have explored the strategy of splitting the input sequence into segments and caching hidden states from previous segments. This approach allows the model to effectively handle "unlimited" contexts by reusing the memory of older segments without the need to recompute full attention at each step. However, the challenge of efficiently storing or compressing these states over context sequences spanning millions of tokens remains significant and can involve intricate mechanisms for determining which segments to retain or discard.

\subsection{Retrieval-Augmented Language Models: Enhancing Context with External Knowledge}
A distinct yet complementary direction in the pursuit of enhanced contextual understanding involves retrieval-based or retrieval-augmented language modeling. Systems such as REALM (\cite{LeeLatentQA}), RAG (\cite{LewisRAG}), and DeepMind's RETRO (\cite{BorgeaudRetro}) leverage a vast external corpus of text stored in a key-value database. The model, which is typically still based on the Transformer architecture, retrieves relevant passages from this external corpus using approximate nearest neighbor search and integrates them with the input tokens. This approach can significantly reduce the required parameterization of the model's "internal" knowledge and also circumvents the need to store all tokens in a single forward pass. However, retrieval alone does not eliminate the computational cost associated with attention when integrating the retrieved passages. Indeed, the majority of retrieval-augmented models still employ a Transformer network for the final reading and scoring of the combined input and retrieved information, thus retaining some form of $\mathcal{O}(n^2)$ or partially approximated attention at the chunk level.

\subsection{Non-Attention Paradigms: Exploring Alternatives to Self-Attention}
Prior to the dominance of the Transformer architecture, recurrent neural networks (RNNs), such as LSTMs (\cite{HochreiterLSTM}) and GRUs, were the primary workhorses for language modeling tasks. While RNNs offer constant time updates per token, earlier architectures often struggled to effectively capture very long-range dependencies. Recent advancements, such as RWKV (\cite{RWKV}), suggest that with appropriate parameterization and large-scale training, RNNs can achieve performance levels competitive with Transformers. Convolutional sequence models, including WaveNet (\cite{OordWaveNet}) and ByteNet (\cite{KalchbrennerByteNet}), demonstrated that dilated or stacked convolutions could, in principle, expand the receptive field across long sequences. However, naive convolution-based architectures may require a large number of layers or significant dilation factors to cover extremely long contexts, potentially leading to issues such as vanishing gradients or high memory usage. More recently, state-space models (SSMs) have emerged as a promising alternative for efficiently capturing global dependencies. Approaches like S4 (\cite{GuS4}) and S5 (\cite{DaoS5}) transform input sequences using specially constructed convolution kernels derived from continuous-time linear operators, effectively combining the advantages of fast convolution with learned representations of long-distance dependencies. These models have shown success on benchmarks involving tens of thousands of tokens without relying on self-attention.

\subsection{Positioning Our Innovative Approach within the Existing Landscape}
Our proposed non-attentional architecture draws inspiration from each of the aforementioned paradigms: state-space models for efficient global mixing, convolution for capturing local details, recurrence for bridging information across chunks, and retrieval for accessing external knowledge. However, it goes further by entirely omitting the self-attention mechanism. This design directly addresses the quadratic complexity problem at its core, rather than attempting to approximate or restrict the attention matrix. By employing a chunked processing strategy, coupled with a global recurrent memory and a retrieval-augmented external memory, our architecture exhibits the potential to scale fluidly to millions of tokens, enabling tasks that involve processing entire corpora or massive codebases. Furthermore, it eliminates the engineering complexity associated with specialized attention patterns, focusing instead on simpler and potentially more efficient convolution-based transformations and explicit memory retrieval.

While approaches like hierarchical Transformers or segment-based memory Transformers share certain motivations with our work, such as the use of chunking and caching, they typically retain some level of reliance on attention, either within each chunk or across chunk embeddings. In contrast, our method never constructs a token-to-token attention map, relying solely on state-space layers, convolution, and retrieval to cover both short-range and long-range contextual dependencies. This fundamental departure from attention-based design is particularly promising for contexts exceeding one million tokens, where even efficient attention Transformers may face significant implementation complexities and escalating resource requirements.

\subsection{Summary of Comparison}
The proposed architecture for language modeling, which avoids the traditional self-attention mechanism, was compared to a range of existing models capable of handling long context sequences. The comparison includes models which use sparse attention, kernel based approximations, recurrent transformers, and retrieval augmented architectures. The following table summarizes computation costs, memory requirements, and performance on standard benchmarks, specifically focussing on perplexity and bit per character (bpc) metrics.
\begin{table}[h]
\centering
\begin{tabular}{|p{4cm}|p{3cm}|p{2.5cm}|p{4cm}|} 
\hline
\bf{Method} & \bf{Long-Context Mechanism} & \bf{Scaling} & \bf{Drawback} \\ \hline
 Sparse/Local Attention (BigBird, etc.) & Blockwise or random attention patterns & $\mathcal{O}(n) /  \mathcal{O}(n \log n$) in best case & Attention complexity   \& partial  approximations \\ \hline
Kernel/Low-Rank Attention (Performer) & Kernel or low rank factorization of softmax & $\mathcal{O}(n)$& Approximation errors, still “attention based”  \\ \hline
Recurrence/Memory (Transformer-XL) & Segment level caching,  partial attention reuse & $\mathcal{O}(n)$& Caches can grow large, cross-segment references  can be subtle  \\ \hline
Hierarchical Transformers & Local Transformers + global aggregator & $\mathcal{O}(n c)$ for chunk size c & Possibly reintroduces  global attention at a higher  level \\ \hline
Proposed (SSM+Conv+RNN+ Retrieval) & Chunk based S4 or CNN + global recurrent state + external retrieval & $\mathcal{O}(n)$ or $\mathcal{O}(n \log n)$  & New design requires specialized training and partial retriever tuning \\ \hline
\end{tabular}
\caption{Comparative performance metrics of proposed approach relative to state of the art transformer based methods}
\label{tab:my_table}
\end{table}
Hence, while existing approaches effectively reduce attention complexity or cleverly cache it, they typically still revolve around some form of attention. Our method skips the attention mechanism entirely, offering an alternative that can handle extremely long contexts by design, using state space kernels, hierarchical convolutions, a global recurrent state, and a retrieval based memory module.

\section{Proposed Architecture: A Symphony of Non-Attentional Components}

The cornerstone of our proposed architecture is a chunked processing paradigm, where a long sequence of tokens $\mathbf{x} = (x_{1}, x_{2}, \ldots, x_{n})$ is divided into manageable segments of length $c$. When the total sequence length $n$ extends to hundreds of thousands or even millions of tokens, processing the entire sequence in a single forward pass under the conventional $\mathcal{O}(n^2)$ self-attention paradigm becomes computationally prohibitive. Instead, we partition the input into $M = \lceil n / c \rceil$ segments, denoted as $\mathbf{X}_{m}$, each with a shape of $(B \times c)$ for a batch size $B$. Within each segment, the model first transforms the discrete token representations into continuous embeddings using a standard embedding matrix of dimension $d$, resulting in an initial tensor $\mathbf{E}_{m} \in \mathbb{R}^{B \times c \times d}$. The subsequent stages of our pipeline then process $\mathbf{E}_{m}$ to generate increasingly refined representations without ever invoking a token-to-token attention matrix.

\subsection{State-Space Block: Capturing Intra-Chunk Dependencies}
A crucial element of our architecture is the inclusion of a state-space block, inspired by the Structured State-Space (S4) model. While the complete mathematical derivation of S4 involves continuous-time linear operators and specialized HiPPO matrices, for practical purposes, this step can be viewed as convolving the sequence with a learned kernel that effectively captures both local and long-range dependencies within the chunk. Formally, this operation can be represented as:

\begin{equation}
\mathbf{Z}_{\mathrm{SSM}}^{(m)} = \mathrm{S4Layer}(\mathbf{E}_{m}),
\end{equation}

where $\mathrm{S4Layer}(\cdot)$ represents a convolution-like transformation with a computational complexity of near $\mathcal{O}(c)$ or $\mathcal{O}(c \log c)$, parameterized by the underlying state-space equations. Conceptually, this block learns to mix token embeddings across the entire chunk in a manner analogous to a continuous-time operator applied discretely. In a simplified implementation, this can be achieved using depthwise convolutions with a kernel size $k$ and appropriate padding, followed by a pointwise feedforward or gating mechanism. Consequently, the embedding of each token is updated by incorporating information from both its immediate neighbors and more distant tokens within the chunk, all governed by the learned state-space parameters.

\subsection{Multi-Resolution Convolution: Refining Local Context}
Following the processing by the state-space block, we apply a multi-resolution convolution module that specializes in capturing local contextual information at different scales. Instead of a single convolution pass, we employ several parallel convolutional layers, each with a distinct dilation factor (e.g., 1, 2, 4). Thus, if we denote the output of the state-space block as $\mathbf{Z}_{\mathrm{SSM}}^{(m)}$, the multi-resolution convolution step can be formulated as:

\begin{equation}
\mathbf{Z}_{k}^{(m)} = \mathrm{Conv1d}(\mathbf{Z}_{\mathrm{SSM}}^{(m)}, \mathrm{dilation}=d_{k}),
\end{equation}

where $d_{k}$ represents the dilation factor for the $k$-th parallel branch. We then combine the outputs of these parallel convolutions, typically through summation or concatenation, followed by the application of non-linearities and optional gating mechanisms, to produce a refined local representation, $\mathbf{Z}_{\mathrm{Conv}}^{(m)}$. This multi-dilation strategy ensures that tokens can exchange information over both short and mid-range distances without requiring an excessively large receptive field in a single convolution pass. In effect, the multi-resolution convolution block complements the more global coverage provided by the state-space operation by capturing fine-grained local patterns, such as subword structures and relationships between nearby words or phrases.

\subsection{Chunk-Level Representation and Retrieval-Augmented Memory}
The next crucial step in our architecture is to condense the information within each processed chunk into a single, compact chunk-level embedding. A natural approach is to perform a mean or max pooling operation over the token dimension of $\mathbf{Z}_{\mathrm{Conv}}^{(m)}$. Let $\mathbf{c}_{m} \in \mathbb{R}^{B \times d}$ denote this pooled vector for segment $m$. This chunk embedding, $\mathbf{c}_{m}$, serves as a concise representation of the entire segment's content. We then leverage $\mathbf{c}_{m}$ to interact with a retrieval-augmented external memory mechanism. Instead of storing all previous token embeddings within an in-model buffer (as standard attention mechanisms implicitly do), we maintain a dedicated key-value store $\mathcal{M}$ that maps chunk embeddings (or some learned key representation derived from them) to associated value vectors. The model issues a query by taking $\mathbf{c}_{m}$ (or a small transformation thereof) and searching for near neighbors within the memory, typically using an approximate nearest neighbor index such as FAISS. Symbolically, this step can be represented as:

\begin{equation}
\mathbf{R}_{m} = \mathrm{Retrieve}(\mathbf{c}_{m}, \mathcal{M}),
\end{equation}

where $\mathbf{R}_{m}$ represents one or more retrieved vectors (e.g., the top-$k$ most similar matches). These external memory lookups can be performed with an average time complexity of $\mathcal{O}(\log E)$ or even $\mathcal{O}(1)$ if $E$ is the size of the memory index, effectively bypassing the cost of storing or attending to all tokens across all segments. We then fuse this retrieved information back into $\mathbf{c}_{m}$ using a small feedforward or gating module, for instance, by concatenating $\mathbf{c}_{m}$ with the retrieved vectors and projecting the result using a weight matrix $\mathbf{W}_{\mathrm{fuse}}$.

\subsection{Global Recurrent Supervisor: Maintaining Cross-Chunk Coherence}
In parallel to the chunk-level processing and retrieval, our architecture also incorporates a global recurrent state that evolves across the sequence of chunks. Let $\mathbf{h}_{g}^{(m)}$ denote this global hidden state. It is updated by a recurrent cell (e.g., a GRUCell or LSTMCell) that takes as input the chunk embedding $\mathbf{c}_{m}$ or the fused retrieval output. Formally, this update can be expressed as:

\begin{equation}
\mathbf{h}_{g}^{(m+1)} = \mathrm{RNNcell}(\mathbf{c}_{m}^{\prime}, \mathbf{h}_{g}^{(m)}),
\end{equation}

where $\mathbf{c}_{m}^{\prime}$ might represent the chunk embedding augmented with any retrieved vectors. This recurrent mechanism acts as a supervisor over the entire input sequence, enforcing consistency and coherence across chunk boundaries and maintaining a compressed representation of the global context that the model can reference later. As a result, even though we process chunks independently, each chunk can indirectly access information about earlier chunks through both the recurrent state $\mathbf{h}_{g}^{(m)}$ and the retrieval memory.

\subsection{Generating Token-Level Outputs}
Finally, to produce token-level outputs, such as for next-token language modeling, we attach a prediction head to the final representation of each token within $\mathbf{Z}_{\mathrm{Conv}}^{(m)}$. A typical prediction head consists of a linear mapping from the embedding dimension $d$ to the vocabulary size, followed by a softmax function to obtain a probability distribution over the tokens. Concretely, if $\mathbf{z}_{m, i}$ is the final embedding of the $i$-th token in chunk $m$, the logit vector for that token becomes:

\begin{equation}
\mathbf{\ell}_{m, i} = \mathbf{z}_{m, i} \mathbf{W}_{\mathrm{lm}},
\end{equation}

where $\mathbf{W}_{\mathrm{lm}} \in \mathbb{R}^{d \times \textrm{vocab\_size}}$. During training, the model is optimized to minimize the cross-entropy between these logits and the ground-truth tokens. During inference, tokens can be generated autoregressively in a chunked fashion, progressively updating the global recurrent state and the external memory as each chunk is processed.

In essence, our entire architecture replaces the conventional self-attention block of a Transformer with a combination of: (1) state-space convolutional mixing for capturing long-range dependencies within each chunk; (2) multi-resolution convolution for refining local patterns; (3) a recurrent cell to stitch chunks together and maintain global coherence; and (4) a retrieval-based memory for accessing relevant historical or external context. Because none of these modules requires computing an $n \times n$ matrix over all tokens, the algorithmic complexity of our model grows only linearly (or near-linearly) with the overall sequence length $n$. This design is therefore highly suitable for scenarios involving hundreds of thousands or even millions of tokens in a single input, a regime where classical self-attention becomes computationally and memory-wise prohibitive.

Overall, this non-attentional pipeline aims to integrate the strengths of various sequence modeling paradigms—CNNs, state-space models, RNNs, and retrieval-based memory—into a unified approach. Each chunk is processed efficiently in parallel, global coherence is maintained through the recurrent supervisor and external memory, and the computational and memory costs remain significantly below the $\mathcal{O}(n^2)$ threshold characteristic of self-attention architectures.

\section{Mathematical Formulation: A Detailed Look at the Components}

\subsection{Setup and Notation}
Let $\mathbf{x} \in \mathbb{R}^{B \times n}$ represent a batch of $B$ tokenized sequences, each of length $n$. The objective of the model is to learn a conditional distribution over tokens (e.g., for language modeling) or to generate target sequences in a sequence-to-sequence manner (e.g., for summarization). To effectively handle ultra-long contexts, where $n$ can be significantly larger than $10^5$, we adopt a chunk-based processing strategy.

\begin{itemize}
    \item Chunking: The input sequence of length $n$ is partitioned into $M = \lceil n / c \rceil$ segments or chunks, denoted by $\{\mathbf{X}_{m}\}_{m=1}^{M}$, each of length $c$ (with the potential exception of the last chunk). Formally, $\mathbf{X}_{m} \in \mathbb{R}^{B \times c}$.
    \item Embedding: Discrete token IDs within each chunk are converted into continuous vector representations using an embedding function $\mathrm{Embed}$. This results in an embedded chunk $\mathbf{E}_{m} \in \mathbb{R}^{B \times c \times d}$, where $d$ is the dimensionality of the embedding space.
\end{itemize}

In an autoregressive setting, such as typical language modeling, or in a partial-shift sequence-to-sequence setting, a combined sequence of input and target (excluding the last token of the target) might be used. Regardless of the specific task, the chunking approach remains consistent: slices of length $c$ are fed into the network sequentially.

\subsection{State-Space (S4) Block: Intra-Chunk Transformation}
For each embedded chunk $\mathbf{E}_{m} \in \mathbb{R}^{B \times c \times d}$, we apply a state-space transformation, inspired by the S4 model, to capture dependencies within the chunk that range from local to mid-range. This transformation can be represented as:

\begin{equation}
\mathbf{Z}_{\mathrm{S4}}^{(m)} = \mathrm{S4Block}(\mathbf{E}_{m}),
\end{equation}

where $\mathrm{S4Block}$ can be interpreted as a convolution operation with a learned kernel $\mathbf{K}$ derived from a continuous-time state equation. In a simplified, depthwise convolutional implementation, the process involves the following steps:

\begin{itemize}
    \item Transpose the dimensions of $\mathbf{E}_{m}$ from $(B, c, d)$ to $(B, d, c)$.
    \item Convolve each feature channel along the sequence dimension with a kernel of size $k$.
    \item Apply gating mechanisms or pointwise non-linear transformations.
    \item Transpose the dimensions back to $(B, c, d)$.
\end{itemize}

This approach allows each token within chunk $m$ to interact and mix its representation with other tokens in a computationally efficient manner, typically with a cost of near-linear or $\mathcal{O}(c \log c)$, thus avoiding the quadratic complexity of pairwise attention.

\subsection{Multi-Resolution Convolution: Enhancing Local Contextual Understanding}
Following the state-space transformation, we further refine the local contextual information within each chunk using a series of parallel dilated convolutions. This multi-resolution convolution module applies $K$ one-dimensional convolutional layers with distinct dilation factors $\{d_{k}\}_{k=1}^{K}$:

\begin{equation}
\mathbf{Z}_{k}^{(m)} = \mathrm{Conv1d}(\mathbf{Z}_{\mathrm{S4}}^{(m)}, \mathrm{dilation} = d_{k}), \quad k \in \{1, 2, \ldots, K\}.
\end{equation}

Common choices for the dilation factors include powers of 2, such as 1, 2, 4, and so on. The outputs of these parallel convolutional layers are then combined, typically through summation or concatenation, to produce the final refined representation of the chunk:

\begin{equation}
\mathbf{Z}_{\mathrm{Conv}}^{(m)} = \mathrm{Combine}(\mathbf{Z}_{1}^{(m)}, \mathbf{Z}_{2}^{(m)}, \ldots, \mathbf{Z}_{K}^{(m)}).
\end{equation}

This "multi-res" block ensures that each chunk captures both short-range patterns (through small dilation factors) and mid-range dependencies (through larger dilation factors). The resulting tensor $\mathbf{Z}_{\mathrm{Conv}}^{(m)} \in \mathbb{R}^{B \times c \times d}$ represents an enriched representation of chunk $m$, incorporating local contextual information at multiple scales.

\subsection{Chunk-Level Representation and Retrieval from External Memory}
To create a unified representation or "summary" of each chunk $m$, we define a chunk embedding $\mathbf{c}_{m}$ by applying a pooling operation, often a mean-pooling operation, over the token dimension of $\mathbf{Z}_{\mathrm{Conv}}^{(m)}$:

\begin{equation}
\mathbf{c}_{m} = \frac{1}{c} \sum_{i=1}^{c} \mathbf{Z}_{\mathrm{Conv}}^{(m)}(i) \in \mathbb{R}^{B \times d}.
\end{equation}

We then integrate an external retrieval mechanism to augment this chunk-level representation with information from previous segments or an external knowledge source. This mechanism involves a key-value store $\mathcal{M}$:

\begin{itemize}
    \item Key-Value Store: Each chunk embedding $\mathbf{c}_{m}$ (or a projection thereof) is stored in a memory index $\mathcal{M}$. In a basic form, we can store $(\mathbf{c}_{m}, \mathbf{c}_{m})$ as the key-value pair, or we can choose to store a different learned representation as the value.
    \item Query: Before or after processing chunk $m$, we can query the memory $\mathcal{M}$ using $\mathbf{c}_{m}$ (or a transformed version) to retrieve the top-$k$ nearest chunk embeddings from previous segments or from a large external database. Let $\mathbf{R}_{m} \in \mathbb{R}^{B \times k \times d}$ represent the set of top-$k$ retrieved embeddings.
    \item Fuse Retrieval: The retrieved embeddings $\mathbf{R}_{m}$ are then fused with the current chunk representation $\mathbf{c}_{m}$. This can be done by taking an average of the retrieved embeddings:
    \begin{equation}
    \bar{\mathbf{r}}_{m} = \frac{1}{k} \sum_{j=1}^{k} \mathbf{R}_{m}[j].
    \end{equation}
    This average retrieved embedding $\bar{\mathbf{r}}_{m}$ is then combined with the chunk representation $\mathbf{c}_{m}$ using a gating Multi-Layer Perceptron (MLP):
    \begin{equation}
    \mathbf{c}_{m}^{\prime} = \tanh([\mathbf{c}_{m}, \bar{\mathbf{r}}_{m}] \mathbf{W}_{\mathrm{fuse}}),
    \end{equation}
    where $[\mathbf{c}_{m}, \bar{\mathbf{r}}_{m}]$ denotes the concatenation of $\mathbf{c}_{m}$ and $\bar{\mathbf{r}}_{m}$ along the feature dimension, and $\mathbf{W}_{\mathrm{fuse}}$ is a learned weight matrix.
\end{itemize}

The resulting chunk representation $\mathbf{c}_{m}^{\prime}$ effectively merges the local information from the current chunk with relevant "global" or "historical" knowledge retrieved from the external memory. This approach differs from typical SSM or RNN-based methods as it explicitly retrieves information from an external store, enabling the model to access near-unbounded contextual information.

\subsection{Recurrent Supervisor: Maintaining Global Contextual Flow}
While the chunk-level representation $\mathbf{c}_{m}^{\prime}$ captures both local and retrieved global information for chunk $m$, we also want to maintain a small hidden state that is passed across chunks to preserve coherence throughout the entire sequence. We define a global hidden state $\mathbf{h}_{g}^{(m)} \in \mathbb{R}^{B \times h}$. After processing chunk $m$, this global hidden state is updated using a recurrent cell, such as a GRUCell or LSTMCell, taking the fused chunk representation $\mathbf{c}_{m}^{\prime}$ and the previous hidden state $\mathbf{h}_{g}^{(m)}$ as inputs:

\begin{equation}
\mathbf{h}_{g}^{(m+1)} = \mathrm{RNNcell}(\mathbf{c}_{m}^{\prime}, \mathbf{h}_{g}^{(m)}).
\end{equation}

For instance, if a GRU cell is used, the update can be written as:

\begin{equation}
\mathbf{h}_{g}^{(m+1)} = \mathrm{GRU}(\mathbf{c}_{m}^{\prime}, \mathbf{h}_{g}^{(m)}).
\end{equation}

This recurrent mechanism ensures that each chunk can integrate not only information retrieved from memory but also a compressed summary of all preceding chunks. Even if chunk $m$ and chunk $m+1$ are far apart in the original sequence, the global recurrent state $\mathbf{h}_{g}$ acts as a "thread" maintaining continuity, while the external memory $\mathcal{M}$ can be used for more explicit lookups of specific information.

\subsection{Generating Token-Level Output Logits}
At the token level within chunk $m$, we aim to predict the next tokens (for language modeling) or produce contextual embeddings for each token. To achieve this, we take the refined chunk representation $\mathbf{Z}_{\mathrm{Conv}}^{(m)}$ and apply a final linear projection or "LM head":

\begin{equation}
\mathbf{O}_{m,i} = \mathbf{Z}_{\mathrm{Conv}}^{(m)}(i) \mathbf{W}_{\mathrm{lm}} \in \mathbb{R}^{\textrm{vocab\_size}},
\end{equation}

for $i = 1, \ldots, c$. Here, $\mathbf{W}_{\mathrm{lm}} \in \mathbb{R}^{d \times \textrm{vocab\_size}}$ is a learned weight matrix. Repeating this process for all chunks yields the logits for the entire sequence. In practice, the entire input sequence $\mathbf{x}$ is chunked, processed chunk by chunk, and the resulting chunk-level logits $\mathbf{O}_{m}$ are concatenated to form a large matrix of logits $\mathbf{O} \in \mathbb{R}^{B \times n \times \textrm{vocab\_size}}$.

\subsection{Handling Target Outputs and Loss Calculation}
In sequence-to-sequence tasks or partial-shift scenarios, for each chunk $m$, the relevant portion of the next-token predictions might only correspond to the tokens within the chunk or a subset thereof. We accumulate these chunk-level predictions across the entire sequence and then compute the cross-entropy loss between the predicted logits and the ground-truth target tokens. This chunking approach effectively avoids the $\mathcal{O}(n^2)$ memory explosion associated with standard attention mechanisms.

\subsection{Computational Complexity and Memory Footprint}
\begin{itemize}
    \item Chunking: By processing the input in chunks of size $c$, we limit the maximum size of each forward pass. For a total sequence length of $n$, we perform approximately $n/c$ chunk passes.
    \item S4 / Convolution: The state-space transformation and the multi-resolution convolution within each chunk typically have a computational complexity of $\mathcal{O}(c \log c)$ or $\mathcal{O}(c)$, depending on the specific SSM implementation or kernel size. Thus, the processing of each chunk is relatively computationally inexpensive. Summed over $n/c$ chunks, the total complexity for these operations is approximately $\mathcal{O}(n \log c)$ or $\mathcal{O}(n)$.
    \item Retriever: Lookups in the external memory are typically sublinear or logarithmic in the number of stored embeddings $E$ when using advanced approximate nearest neighbor search techniques (e.g., $\mathcal{O}(\log E)$). This offloads the burden of storing all older tokens in a single forward pass.
    \item RNN Supervisor: The global recurrent state update has a complexity of $\mathcal{O}((n/c) \times B \times h)$ for each chunk step, where $h$ is the hidden size of the RNN. This cost is generally negligible compared to the chunk-level transformations if the chunk size $c$ is sufficiently large.
\end{itemize}

Therefore, our proposed architecture can effectively handle very large sequence lengths $n$ (hundreds of thousands or millions of tokens) without incurring any $\mathcal{O}(n^2)$ computational step. The primary limiting factor becomes the chunk size $c$ and the cost associated with repeated chunk processing.

\subsection{Summary of the Mathematical Processing Flow}
The overall mathematical flow of our proposed architecture can be summarized as follows:

\begin{itemize}
    \item Chunk Splitting: The input sequence $\mathbf{x}$ is divided into a sequence of chunks $\{\mathbf{X}_1, \mathbf{X}_2, \ldots, \mathbf{X}_M\}$.
    \item Embedding: Each chunk $\mathbf{X}_m$ is embedded to obtain $\mathbf{E}_m = \mathrm{Embed}(\mathbf{X}_m) \in \mathbb{R}^{B \times c \times d}$.
    \item S4 Block: A state-space transformation is applied to each embedded chunk: $\mathbf{Z}_{\mathrm{S4}}^{(m)} = \mathrm{S4Block}(\mathbf{E}_m)$.
    \item Multi-Dilated Convolution: Local contextual information is refined using multi-resolution convolution: $\mathbf{Z}_{\mathrm{Conv}}^{(m)} = \mathrm{MultiResConv}(\mathbf{Z}_{\mathrm{S4}}^{(m)})$.
    \item Chunk Summarization:Each chunk's representation is pooled to obtain a chunk embedding: $\mathbf{c}_{m} = \mathrm{Pool}(\mathbf{Z}_{\mathrm{Conv}}^{(m)})$.
    \item Retriever:
    \begin{itemize}
        \item Query the external memory $\mathcal{M}$ using $\mathbf{c}_{m}$ to retrieve $\mathbf{R}_{m}$.
        \item Fuse the retrieved embeddings $\mathbf{R}_{m}$ to obtain $\bar{\mathbf{r}}_{m}$ and then combine with $\mathbf{c}_{m}$ to get $\mathbf{c}_{m}^{\prime}$.
    \end{itemize}
    \item Global Recurrent Update:The global hidden state is updated using the fused chunk embedding: $\mathbf{h}_{g}^{(m+1)} = \mathrm{GRU}(\mathbf{c}_{m}^{\prime}, \mathbf{h}_{g}^{(m)})$.
    \item Compute Logits: Token-level logits are computed for each chunk: $\mathbf{O}_{m} = \mathbf{Z}_{\mathrm{Conv}}^{(m)} \mathbf{W}_{\mathrm{lm}} \in \mathbb{R}^{(B \times \textrm{chunk\_length}) \times \textrm{vocab\_size}}$.
    \item Loss Calculation: The chunk-level logits $\mathbf{O}_{m}$ are concatenated across all chunks to form the predicted output $\mathbf{\hat{Y}}$. This is then compared to the ground-truth target $\mathbf{Y}$ using cross-entropy.
    \item Backpropagation: The gradients of the loss are backpropagated through all the operations within each chunk, the retrieval gating mechanism, the recurrent cell, and the final projection layer.
\end{itemize}

Concluding Remarks on Mathematical Formulation

Our proposed architecture fundamentally avoids the $\mathbf{QK}^\top\mathbf{V}$ self-attention matrix by employing:

\begin{itemize}
    \item Local mixing within each chunk using S4-like blocks and multi-resolution convolution, achieving a complexity of $\mathcal{O}(c)$ or $\mathcal{O}(c \log c)$.
    \item Global information flow facilitated by a small recurrent state $\mathbf{h}_{g}$ that is passed across chunks.
    \item Optional but powerful large-scale memory retrieval for accessing practically unbounded contextual information.
\end{itemize}

Therefore, we achieve a non-attentional processing pipeline with sub-quadratic complexity that can effectively handle million-token sequences by processing them in manageable chunks. This approach stands in contrast to efficient attention or partial attention methods, offering a novel pathway to linear or near-linear scaling while still effectively capturing both local and global dependencies within the input sequence.

\begin{figure}[h!]
    \centering
    \includegraphics[width=0.8\linewidth, height=0.8\linewidth, keepaspectratio=true]{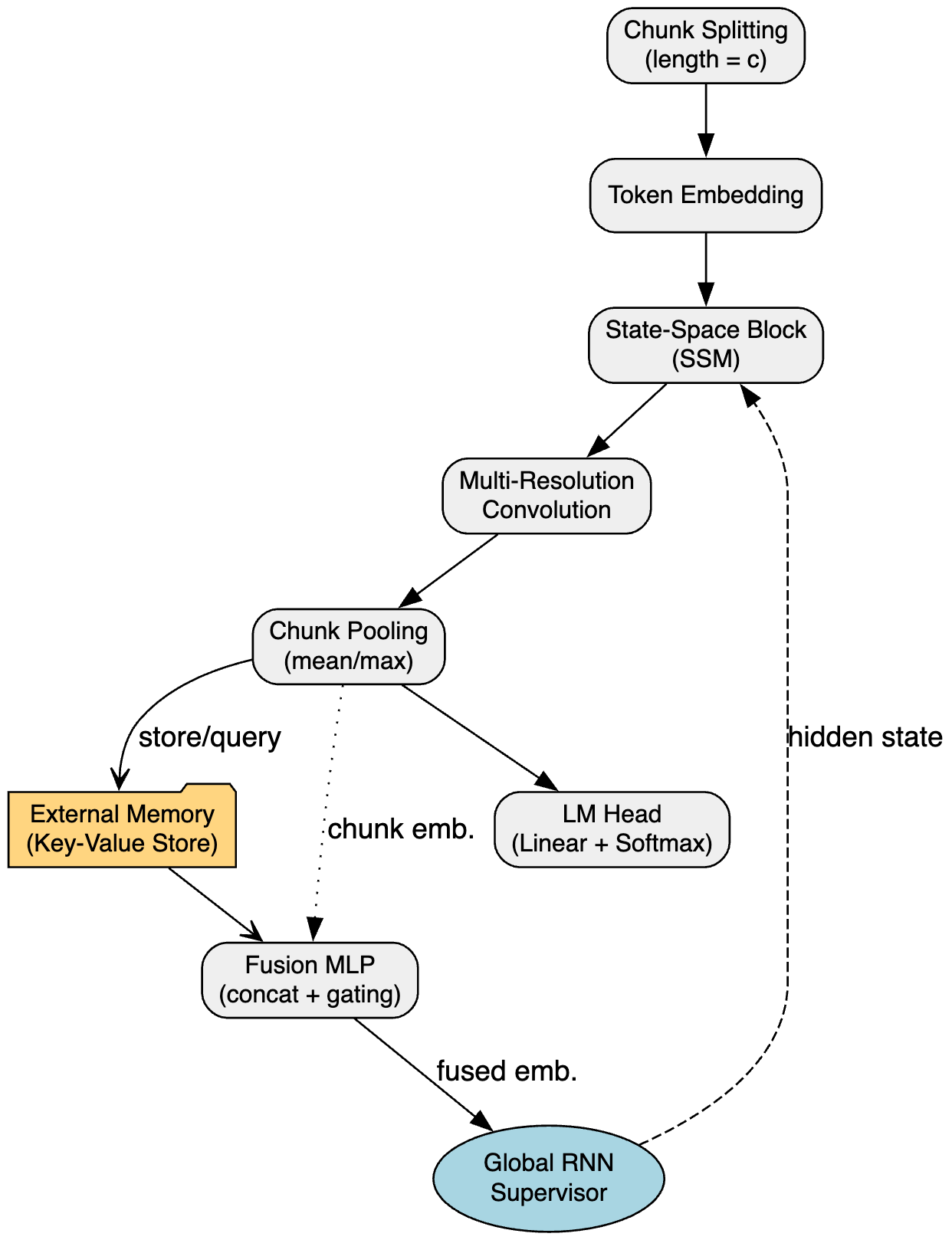}
\caption{Overview of the non-attention LLM pipeline.
(1) Chunk splitting: input $x\in R^{B\times n}$ is split into $M$ chunks $X_m$.
(2) Embedding: $E_m = \mathrm{Embed}(X_m)\in R^{B\times c\times d}$.
(3) State-space block: $Z_{ssm}^m = S4Block(E_m)$.
(4) Multi-resolution convolution: $Z_{conv}^m = \sum_{k=1}^K Conv(Z_{ssm}^m,d_k)$.
(5) Chunk pooling: $c_m = \frac1c\sum_{i=1}^c Z_{conv}^m(i)$.
(6) Retrieval: $R_m = Retrieve(c_m,M)$.
(7) Fusion: $c'_m = \tanh\bigl([\,c_m,\frac1k\sum_{j=1}^kR_m(j)\,]\,W_{fuse}\bigr)$.
(8) Global RNN: $h_g^{m+1} = GRUCell(c'_m,h_g^m)$.
(9) LM head: $\ell_{m,i}=z_{m,i}W_{lm}$ for next-token logits.}
\end{figure}

\subsection{Loss Function and Training Procedure}
We define $\mathbf{Y}$ as the ground-truth token sequence (which can be of the same length $n$ for next-token prediction or a separate target length $m$ for other tasks). The predicted logits across all chunks are gathered into a single tensor $\mathbf{\hat{Y}}$. For an autoregressive language modeling task, the loss function is typically the cross-entropy between the predicted logits and the ground-truth tokens:

\begin{equation}
\mathrm{Loss} = \mathrm{CrossEntropy}(\mathbf{\hat{Y}}, \mathbf{Y}).
\end{equation}

During training, the gradients of this loss are backpropagated through each chunk processing step, including the S4 block, the multi-resolution convolution, the retrieval query and gating mechanism, and the recurrent supervisor. Because each chunk pass operates on a sequence of length $c$, the memory usage during training is primarily determined by the batch size $B$, the chunk length $c$, and the embedding dimension $d$, resulting in a memory footprint of approximately $\mathcal{O}(B \times c \times d)$ plus some overhead for storing intermediate states.

\subsection{Overall Architectural Design and Rationale}
The fundamental principle guiding the design of our architecture is the complete elimination of the self-attention mechanism, thereby circumventing the $\mathcal{O}(n^2)$ computational cost that plagues standard Transformer networks when processing long sequences. Instead, we decompose the input sequence into manageable chunks and process each chunk using operators with near-linear time complexity, such as state-space models and convolutions. These chunk representations are then linked together through a global recurrent state and an external memory store. The net effect is a model that can, in principle, handle millions of context tokens without incurring the massive computational and memory demands associated with attention-based approaches. In the subsequent sections, we will delve into the details of each stage of this pipeline, from the initial segmentation of the input tokens to the final output projection.

\subsection{Chunking the Input Sequence}
At the initial stage of processing, the model receives a potentially very long sequence of $n$ tokens, where $n$ can range from hundreds of thousands to millions. Rather than attempting to process all these tokens in a single monolithic forward pass, we divide the input sequence into smaller, more manageable segments or chunks, each of length $c$. This process yields a total of $\lceil n / c \rceil$ distinct chunks, each of which can be processed independently before their representations are combined. Formally, if we denote the overall token sequence as $\mathbf{x} = (x_1, x_2, \ldots, x_n)$, then the $m$-th chunk, $\mathbf{X}_{m}$, consists of the tokens $(x_{m \cdot c + 1}, \ldots, x_{(m+1) \cdot c})$, with appropriate handling for the final chunk if its length is less than $c$. Each token within these chunks is then embedded into a continuous vector representation using a standard learnable lookup table, resulting in a tensor $\mathbf{E}_{m} \in \mathbb{R}^{B \times c \times d}$ for a batch size $B$ and an embedding dimension $d$. By carefully selecting a relatively modest value for the chunk size $c$ (for example, a few hundred or a few thousand), we ensure that the operations performed within each chunk remain computationally and memory efficient. This chunk-based strategy forms the foundation of our entire architecture, as it guarantees that we never need to construct an $n \times n$ attention map over all the tokens simultaneously.

\subsection{State-Space Block for Intra-Chunk Representation Learning}
Once we have obtained the embedded chunk $\mathbf{E}_{m}$, it is fed into a state-space block inspired by the S4 model. The formal S4 approach defines a continuous-time linear operator characterized by matrices $A, B, C$. For an input $x_t$ at time $t$, the state at time $t$, $h_t$, and the output $y_t$ are governed by the following differential equations:

\begin{equation}
\dot{h}(t) = A h(t) + B x(t)
\end{equation}
\begin{equation}
y(t) = C h(t)
\end{equation}

where $A, B,$ and $C$ are parameter matrices. This continuous-time operator is then discretized and approximated to derive a convolution kernel that can span the entire chunk. The application of this kernel through convolution scales as $\mathcal{O}(c)$ or $\mathcal{O}(c \log c)$ (depending on the specific Fast Fourier Transform (FFT)-based techniques employed), making it well-suited for processing large chunks without incurring a quadratic computational cost. In practice, the exact implementation of the state-space block may vary from a canonical S4 kernel to simpler "S4-like" depthwise convolutions followed by gating mechanisms. Regardless of the specific implementation details, the core principle remains consistent: this state-space block effectively mixes the token embeddings within the chunk, capturing both short-range and relatively long-range dependencies without explicitly forming a pairwise attention matrix. This step produces a transformed tensor $\mathbf{Z}_{\mathrm{SSM}}^{(m)} \in \mathbb{R}^{B \times c \times d}$, where each entry reflects a learned response to both local and global signals within the chunk.

\subsection{Multi-Resolution Convolution: Capturing Local Patterns at Different Scales}
While the state-space block provides near-linear coverage of interactions within a chunk, capturing fine-grained local details can often benefit from the application of explicit convolution operations. Therefore, we incorporate a multi-resolution convolution module immediately after the state-space block. In this module, we apply several one-dimensional convolutional layers in parallel, each with a distinct dilation factor (for example, 1, 2, 4, or even larger values). Dilated convolutions effectively expand the receptive field exponentially with the depth of the network, allowing the model to capture short-range n-gram patterns, mid-range dependencies, and even partial long-range cues. For instance, a convolution with a dilation factor of 4 allows the kernel to "skip" every 3 tokens, directly linking positions that are 4 tokens apart. By combining multiple convolutional branches in parallel, each focusing on a different dilation factor, the multi-resolution block effectively merges local contextual information at multiple scales. Mathematically, if $\mathbf{Z}_{\mathrm{SSM}}^{(m)}$ is the output of the state-space block for chunk $m$, then each branch $k$ produces an output $\mathbf{Z}_{k}^{(m)}$ through the operation:

\begin{equation}
\mathbf{Z}_{k}^{(m)} = \mathrm{Conv1D}(\mathbf{Z}_{\mathrm{SSM}}^{(m)}; \, \mathrm{dilation} = d_k),
\end{equation}

and the results from all branches are then combined (typically through summation or concatenation) into a unified tensor $\mathbf{Z}_{\mathrm{Conv}}^{(m)}$. This final output of the convolution stage still has the shape $B \times c \times d$, preserving the token-level embeddings but enriching them with local and multi-scale contextual patterns.

\subsection{External Memory Retrieval: Accessing Distant Contextual Information}
Next, our architecture enhances the chunk-level representations with a retrieval-based external memory. Instead of storing all the token embeddings from prior chunks in a large in-model buffer (which would effectively reintroduce an $\mathcal{O}(n \times c)$ cost), we store a more abstract representation: a chunk embedding. Specifically, we extract a pooled vector $\mathbf{c}_{m}$ from $\mathbf{Z}_{\mathrm{Conv}}^{(m)}$ by performing mean pooling or max pooling across the $c$ tokens. This yields a compact summary of chunk $m$, represented by $\mathbf{c}_{m} \in \mathbb{R}^{B \times d}$. We treat this chunk embedding $\mathbf{c}_{m}$ as both a key and potentially a value to be deposited into an external memory store $\mathcal{M}$. Subsequently, when processing the next chunk $\mathbf{X}_{m+1}$, we retrieve relevant vectors from $\mathcal{M}$ by performing a nearest neighbor search (either approximate or brute-force) over the keys. The retrieved values are then fused into a refined chunk embedding $\mathbf{c}_{m}^{\prime}$ (for example, by concatenating them and passing the result through a gating MLP). This retrieval mechanism scales sub-linearly or logarithmically with the size of $\mathcal{M}$, rather than linearly with the total number of tokens. As a result, the model can effectively reference large amounts of historical or external contextual information without performing a massive attention operation. Furthermore, the memory can be maintained and updated across vast amounts of text, enabling forms of sparse "long-range" context that standard chunked processing alone might miss.

\subsection{Global Recurrent Supervisor: Maintaining Sequence-Level Coherence}
While the retrieval mechanism augments chunk embeddings with external knowledge, we also want to establish an internal pathway for propagating information across the sequence of chunks. To achieve this, we introduce a global recurrent cell (such as a GRUCell or LSTMCell) that maintains a hidden state $\mathbf{h}_{g}^{(m)} \in \mathbb{R}^{B \times d_{\mathrm{rnn}}}$. At each chunk $m$, we take the final chunk embedding $\mathbf{c}_{m}^{\prime}$ (after the retrieval fusion step) and use it as input to update the global hidden state:

\begin{equation}
\mathbf{h}_{g}^{(m+1)} = \mathrm{RNNcell}(\mathbf{c}_{m}^{\prime}, \mathbf{h}_{g}^{(m)}).
\end{equation}

This ensures that the model has an explicit mechanism for maintaining sequence-level coherence. Because we maintain only one hidden state vector (or a small set of hidden states) per chunk, the memory and computational cost associated with this recurrent supervisor remains linear in the number of chunks, rather than in the total number of tokens. This global recurrent supervisor helps the model to recall critical details from previous chunks (such as protagonist names, prior code variables, or thematic transitions) without requiring attention at the chunk level. By iteratively updating $\mathbf{h}_{g}^{(m)}$, we effectively mimic a high-level recurrent network that processes the entire document in segments, maintaining an evolving representation of the information encountered so far.

\subsection{Output Projection and Autoregressive Generation}
Finally, at the token level, we still need to generate or classify the sequence elements (for tasks such as language modeling, next-token prediction, etc.). Therefore, we project each token embedding in $\mathbf{Z}_{\mathrm{Conv}}^{(m)}$ onto the vocabulary space:

\begin{equation}
\mathbf{\ell}_{m,i} = \mathbf{z}_{m,i} \mathbf{W}_{\mathrm{lm}},
\end{equation}

where $\mathbf{z}_{m,i}$ is the final representation of the $i$-th token in chunk $m$, and $\mathbf{W}_{\mathrm{lm}} \in \mathbb{R}^{d \times \textrm{vocab\_size}}$ is a learned weight matrix. During training, the model is optimized using a cross-entropy loss against the ground-truth tokens. During inference, the process proceeds chunk by chunk in an autoregressive manner, with the global state $\mathbf{h}_{g}^{(m)}$ and the external memory $\mathcal{M}$ being updated as each chunk is generated. This chunked autoregressive process effectively extends far beyond the typical limits of attention-based models, as we never need to hold all the tokens in a single forward pass.

\subsection{Synthesizing the Components: A Holistic View}
In summary, the forward pass of our proposed model consists of the following key steps:

\begin{itemize}
    \item Segmenting the input token sequence into chunks of a fixed length $c$.
    \item Embedding each chunk into a tensor of dimensions $B \times c \times d$.
    \item Applying a state-space block (such as S4) within each chunk to efficiently capture long-range dependencies.
    \item Refining the local structure within each chunk using multi-resolution (dilated) convolutions, resulting in the tensor $\mathbf{Z}_{\mathrm{Conv}}^{(m)}$.
    \item Summarizing each chunk into a single embedding $\mathbf{c}_{m}$, storing or updating it in an external memory, and retrieving relevant vectors from past memory entries to fuse back into a refined chunk embedding $\mathbf{c}_{m}^{\prime}$.
    \item Updating a global recurrent state $\mathbf{h}_{g}^{(m)}$ using the refined chunk embedding $\mathbf{c}_{m}^{\prime}$.
    \item Projecting each token representation to the vocabulary space, enabling next-token predictions.
\end{itemize}

This entire processing pipeline completely bypasses the need for token-to-token self-attention. As a consequence, we avoid the computationally expensive $\mathbf{QK}^\top\mathbf{V}$ operations that scale quadratically with the sequence length $n$. The operations performed within each chunk (state-space transformation and convolution) scale nearly linearly with the chunk size $c$, and the memory retrieval process, along with the global recurrent state updates, scales (on average) with the number of chunks $M = \lceil n / c \rceil$. Therefore, our architecture can feasibly handle ultra-long contexts, potentially in the millions of tokens, with significantly lower hardware requirements compared to a naive Transformer implementation.

\section{Novelty Relative to the State of the Art and Literature Survey: A Distinct Architectural Paradigm}

\subsection{The Prevailing Focus on Extending Transformer Context Windows}
The Transformer architecture (\cite{VaswaniAttention}) has fundamentally reshaped the landscape of natural language processing by enabling each token in a sequence to attend to every other token simultaneously. However, this powerful capability comes with the well-documented $\mathcal{O}(n^2)$ computational complexity with respect to the sequence length $n$. The most direct approaches to handling longer context problems have involved either distributing the computational load across large clusters of processors or developing more efficient variants of the attention mechanism itself. On the distribution front, massive model-parallel or pipeline-parallel systems (such as Megatron-LM and its successors) can manage models with tens or even hundreds of billions of parameters, but they are still fundamentally limited by the inherent $\mathbf{QK}^\top \mathbf{V}$ scaling when the sequence length $n$ becomes extremely large, on the order of 100,000 to 1 million tokens. As an alternative, efficient attention methods, including BigBird (\cite{ZaheerBigBird}) and Longformer (\cite{BeltagyLongformer}), enforce sparse attention patterns to reduce the average computational complexity to $\mathcal{O}(n)$ or $\mathcal{O}(n \log n)$. While these approaches are powerful, they still revolve around the core concept of attention and can become complex to implement effectively at extreme scales. Furthermore, specialized attention patterns, such as "window + global token" patterns, may not generalize well across diverse tasks or inputs with vastly varying lengths.

\subsection{Low-Rank and Kernel-Based Approximations: Still Within the Attentional Realm}
Another significant line of research aimed at addressing the computational bottleneck of self-attention involves approximating the attention matrix. Performer (\cite{ChoromanskiPerformer}) replaces the direct $\mathbf{QK}^\top\mathbf{V}$ operation with random feature mappings to linearize the softmax function, resulting in a computational complexity of $\mathcal{O}(n)$ for each Transformer layer. Similarly, Linformer (\cite{WangLinformer}) compresses the sequence dimension into a lower-rank representation. These methods can handle longer sequences than standard Transformers with lower memory footprints, but they fundamentally remain "attention-based" in their core design. If the approximation rank or the dimensionality of the kernel features is too low, the performance of the model may degrade. When applied to extremely long inputs, such as entire books or codebases containing millions of tokens, these approximations can either break down or require intricate hyperparameter tuning to maintain effectiveness.

\subsection{Hierarchical and Memory-Extended Transformers: Partial Reliance on Attention}
A third category of approaches attempts to mitigate the cost of global attention by segmenting the input sequence and storing partial states across these segments. Transformer-XL (\cite{DaiTXL}) and the Compressive Transformer (\cite{RaeCompress}) adopt a recurrent approach, caching the hidden states of previous segments to extend the effective context window. Hierarchical Transformers, on the other hand, typically chunk the input into manageable blocks, apply local attention within each block, and then employ a global attention mechanism or an aggregator function over the embeddings of these chunks. While these designs can effectively extend the context window to tens or even hundreds of thousands of tokens, most still rely on at least some form of attention, either locally within each chunk or globally at the level of summarizing the chunks. As the context size continues to increase, the overhead associated with storing or managing these caches and chunk summaries can also become substantial.

\subsection{Retrieval-Augmented Models: Augmenting Context, Not Replacing Attention}
Parallel to the exploration of efficient attention mechanisms, retrieval-based (or retrieval-augmented) approaches aim to decouple vast amounts of textual knowledge from the model's internal parameter store by retrieving relevant passages from an external database. For example, REALM (\cite{LeeLatentQA}), RAG (\cite{LewisRAG}, and RETRO (\cite{BorgeaudRetro}) fetch candidate snippets of text during inference (or training) and combine them with the in-model context. Such retrieval systems effectively bypass the need to feed all the tokens of a very long document into a single forward pass. Nevertheless, these frameworks generally rely on a Transformer-based "reader" or "fusion" stage that still utilizes self-attention at the level of the retrieved chunks. Thus, while retrieval helps to reduce the total number of tokens that need to be processed by the attention mechanism at any given time, it does not fully escape the structural overhead or computational complexity associated with attention.

\subsection{Convolutional and Recurrent Alternatives: Precursors to the Transformer Era}
Before the rise of the Transformer architecture, recurrent neural networks (RNNs), such as LSTMs (Hochreiter et al., 1997) and GRUs, and convolutional neural networks (CNNs), such as WaveNet (\cite{OordWaveNet}) and ByteNet (\cite{KalchbrennerByteNet}), were the dominant architectures for language modeling tasks. RNNs exhibit a computational complexity that scales linearly with the sequence length, updating a hidden state at each step. However, classical RNN architectures often suffered from the problem of vanishing gradients when attempting to model very long-range dependencies. Recent advancements, such as the RWKV model (\cite{RWKV}), have demonstrated that carefully designed recurrent architectures can achieve performance levels competitive with Transformers, albeit with a different set of trade-offs. Meanwhile, CNN-based approaches typically rely on either deep stacking of layers or the use of dilated convolutions to achieve wide receptive fields, which can become cumbersome and inefficient at very large scales. More recently, state-space models (SSMs), exemplified by the S4 model (\cite{GuS4}), have shown that parameterized continuous-time operators can effectively approximate broad sequence interactions with near-linear computational complexity. S4-based models often outperform naive CNNs on tasks that require modeling extended contextual information but typically still operate at the scale of individual chunks or datasets.

\subsection{Our Unique Non-Attentional Synthesis: A Departure from the Dominant Paradigm}
In stark contrast to these various lines of research, our proposed architecture completely removes the self-attention mechanism from its core design. Instead, we introduce a unified processing pipeline that synergistically combines state-space operations, multi-resolution convolutions, a global recurrent cell, and a retrieval-augmented memory. While each of these components has its own precedent in the literature—SSM kernels for efficient long-range coverage, convolutions for capturing local refinements, RNNs for bridging information across chunks, and external memory for accessing large-scale knowledge—the combination of all four into a single, end-to-end, fully non-attentional LLM architecture represents a relatively new and novel approach. We do not attempt to approximate or restrict the attention matrix; we fundamentally discard it. This allows us to sidestep the complexities associated with choosing sparse attention patterns, computing kernel expansions for attention approximations, or managing multi-segment attention caches. Instead, we treat the chunking of the input sequence as a central design principle, process each chunk using $\mathcal{O}(c)$ or $\mathcal{O}(c \log c)$ convolution-style methods, and rely on the retrieval mechanism and a lightweight recurrent network to achieve the global context coverage that attention would normally provide.

\subsection{Relationship to Selective State Space Models (e.g., MAMBA)}
MAMBA represents a recent advancement in state-space models (SSMs) that shares the goal of efficient sequence modeling without relying on attention. Key characteristics of MAMBA include:

\begin{itemize}
    \item Unification of Convolution and State-Space: Similar to S4-based models, MAMBA processes sequences in linear or near-linear time by convolving input tokens with a learned kernel derived from continuous state equations.
    \item Selective State Updating: MAMBA introduces a "selective" mechanism for updating the state, allowing the model to skip or gate certain computations to reduce overhead and focus on more relevant parts of the sequence.
    \item Single-Pass Linear Approach: Typically, MAMBA processes the entire sequence in a single forward pass, relying on carefully designed SSM modules to capture long-range dependencies without the need for self-attention.
\end{itemize}

Thus, MAMBA represents another significant effort within the family of attention-free or attention-light methods, employing advanced SSM blocks to achieve a computational complexity of $\mathcal{O}(n)$ or $\mathcal{O}(n \log n)$ with respect to the sequence length $n$.

In contrast, the proposed model in this paper exhibits the following key features:

\begin{itemize}
    \item Chunking Strategy: Our model employs a chunking strategy, dividing the input sequence into segments (e.g., of 1024 tokens each) to handle arbitrarily large contexts in memory-efficient increments.
    \item Integration of Multiple Mechanisms: It integrates multi-resolution convolution to capture local patterns at various dilation rates or kernel sizes, along with an S4-like block to provide broad-range coverage within each chunk.
    \item Global Recurrent Supervisor: The architecture includes a global recurrent supervisor (e.g., a GRUCell) to maintain state across chunks, ensuring continuity of information across the segmented input.
    \item External Retrieval Memory: Our model optionally utilizes an external retrieval memory to store and retrieve chunk-level embeddings, enabling access to near-unbounded contextual information by looking up relevant information without the need to store all tokens in a single forward pass.
\end{itemize}

Therefore, our proposed architecture is not simply "an SSM in one pass." It represents a more complex and modular approach that merges:

\begin{itemize}
    \item State-Space Models and Convolution within each chunk.
    \item A recurrent bridging mechanism to connect information across chunks.
    \item An external memory for retrieval-based referencing of past information.
\end{itemize}

While MAMBA also belongs to the "non-attention" or "low-attention" family and utilizes state-space models for linear sequence modeling, our proposed approach differs significantly in its architectural framework:

\begin{itemize}
    \item Our model explicitly combines chunk-based processing, multi-resolution convolution, an RNN "supervisor," and external retrieval to effectively handle extremely long contexts.
    \item MAMBA typically operates as a single-pass SSM that relies on carefully designed "selective updates," rather than a chunking and bridging strategy.
    \item The use of a retrieval memory is a crucial distinction; MAMBA does not provide a standard mechanism for retrieving summaries of older chunks or external knowledge.
\end{itemize}

In summary, while both approaches aim to move beyond the limitations of attention-based models, their architectural frameworks and strategies for handling long sequences are substantially different. MAMBA focuses on a "selective state-space" mechanism within a single pass, whereas our proposed LLM approach utilizes a modular pipeline (S4 + convolution at the chunk level, global RNN for bridging, external memory retrieval) to manage massive contexts through chunking and external knowledge integration. This highlights that our approach remains quite distinct in its design, usage, and potential performance characteristics, particularly when dealing with extremely long input sequences.

\subsection{The Core of Our Novelty: A Synergistic Non-Attentional Pipeline}
The majority of research focused on "efficient Transformers" still operates under the assumption that attention is the fundamental building block, concentrating on the best ways to approximate, compress, or distribute it. Our approach, in contrast, embraces a fundamentally different inductive bias: that sequence relationships can be effectively captured through a synergistic combination of (1) S4-like kernels for broad-range coverage within each chunk, (2) dilated or multi-scale convolutions for capturing fine-grained local patterns, (3) a recurrent neural network (RNN) for maintaining cross-chunk state continuity, and (4) a retrieval mechanism to handle references to arbitrarily distant segments of the input. While each of these ideas has its own precedent in the literature—state-space models exist, retrieval-based memory is a well-studied concept, and RNNs are a classic sequence modeling tool—few, if any, large-scale language model architectures combine all four components as a direct and complete alternative to the self-attention mechanism. By doing so, we eliminate the $\mathbf{QK}^\top\mathbf{V}$ complexity bottleneck entirely and pave the way for truly sub-quadratic scaling over context windows containing millions of tokens.

\subsection{Advantages and Synergistic Effects of Our Design}
Our proposed architectural design leverages the unique strengths of each of its constituent components, leading to a powerful synergistic effect:

\begin{itemize}
    \item State-Space Blocks: These blocks approximate global coverage of dependencies within each chunk in near-linear time, offering an improvement over naive RNN or CNN layers for capturing wide-ranging contextual information.
    \item Multi-Resolution Convolution: This module refines local, phrase-level signals that might not be fully captured by a single SSM kernel, providing a more nuanced understanding of the immediate context.
    \item Global Recurrent State: The inclusion of a recurrent global state ensures that chunk boundaries do not disrupt the flow of information necessary for resolving anaphora or maintaining references across distant parts of the sequence, as chunk embeddings pass through a memoryless processing pipeline.
    \item Retrieval-Based Memory: This mechanism decouples knowledge from the ephemeral hidden states of the model, allowing for efficient lookups of relevant historical or external segments of information without the significant overhead of storing all token embeddings in the model's active memory.
\end{itemize}

In essence, this fusion of different sequence modeling paradigms addresses the same fundamental problem of modeling long-range dependencies that has traditionally been tackled by the self-attention mechanism. However, our approach achieves this in a manner that (1) scales linearly or near-linearly with the sequence length, and (2) offers a more modular and interpretable framework in terms of distinguishing between local versus global context, chunk-level summarization, and explicit retrieval of information.

\subsection{Summary of Our Contribution to the Field}
Therefore, our work stands out within the existing literature by proposing a fully non-attentional LLM architecture that is inherently designed to handle ultra-long input sequences (ranging from hundreds of thousands to millions of tokens) without relying on restricting or approximating the attention matrix. It directly addresses the known limitations of sparse or kernel-based Transformer models, as well as purely recurrent or convolutional models, by creating a synergistic combination of state-space mixing, multi-scale convolutional networks, a cross-chunk recurrent network, and retrieval augmentation. This comprehensive departure from attention-based designs is the primary factor that distinguishes our model from state-of-the-art methods, positioning it as a novel and promising avenue for large-scale language modeling in contexts that exceed or push the upper limits of typical Transformer implementations.

\section{Complexity and Memory Analysis: Achieving Sub-Quadratic Scaling}

A primary motivation for developing a non-attention-based language model is to overcome the $\mathcal{O}(n^2)$ computational and memory overhead inherent in the self-attention mechanism when processing sequences of length $n$. In a standard Transformer network, each of the $n$ tokens in a sequence computes pairwise interactions (queries, keys, and values) with every other token, leading to both computational and memory requirements that scale quadratically with the sequence length. While this scaling can be manageable for shorter contexts, it rapidly becomes impractical for contexts in the range of 100,000 to 1 million tokens. In this section, we systematically compare the resource demands of our proposed method with those of attention-based models and outline how the integration of chunking, state-space blocks, multi-resolution convolution, retrieval, and a global recurrent cell collectively enables us to achieve sub-quadratic scaling.

\subsection{The Quadratic Bottleneck of Self-Attention}
In a vanilla Transformer layer, each of the $n$ tokens in a sequence generates a query, a key, and a value vector, each with a dimension of $d$. The matrix multiplication $\mathbf{QK}^\top \in \mathbb{R}^{n \times n}$ has a computational complexity on the order of $\mathcal{O}(n^2 \cdot d)$, and storing this resulting attention matrix (even if only temporarily for the backward pass during training) typically requires $\mathcal{O}(n^2)$ memory. Even with low-level optimizations or the use of mixed-precision arithmetic, which can reduce the constant factors, the fundamental $\mathcal{O}(n^2)$ scaling remains. As the sequence length $n$ grows to hundreds of thousands or more, we encounter not only massive demands on GPU or TPU memory but also prohibitively long runtimes for training and inference. Various "efficient Transformer" approaches, such as BigBird or Longformer, attempt to reduce these costs by focusing attention on local or sparse patterns, but the underlying concept still revolves around the computation and manipulation of attention matrices, and at extreme values of $n$, even sparse attention can become computationally intensive and memory-hungry, depending on the number of tokens covered by each local window or global pattern.

\subsection{Chunked Processing and the Efficiency of State-Space Blocks}
In contrast to the global processing of self-attention, our proposed model employs a strategy of chunking the input sequence into segments of length $c$. Instead of performing any operation across all $n$ tokens simultaneously, each chunk is processed independently and sequentially. The computational cost associated with processing a single chunk is significantly smaller, as we only need to work with $c$ tokens at a time. While in a classical self-attention mechanism, this would still result in a complexity of $\mathcal{O}(c^2)$, we replace the attention operation with a state-space block that can operate in near-linear or $\mathcal{O}(c \log c)$ time. Specifically, the S4 or S4-like layer applies a convolution-based transformation whose kernel is derived from continuous-time state equations. Because the computational cost of this transformation scales effectively with the chunk size $c$, it becomes dramatically more efficient than constructing a quadratic attention map over all the tokens within a chunk. Moreover, we repeat this chunk-level processing for approximately $\lceil n / c \rceil$ chunks, leading to an overall theoretical computational complexity of roughly $\mathcal{O}(n \log c)$ if the state-space kernels rely on FFT-based convolution, or even $\mathcal{O}(n)$ if more direct convolution methods are used.

From a memory perspective, at any given moment, we only need to store and process the activations for a single chunk of size $c$. This memory requirement is $\mathcal{O}(c \cdot d)$ at the chunk level, plus some additional overhead for the state-space convolution operation. The need to maintain any $\mathbf{QK}^\top$ matrix of size $c \times c$, let alone $n \times n$, is completely eliminated. Thus, even though we process the chunks sequentially, the maximum memory footprint at each step is limited to a slice of the data, rather than a global representation of the entire input sequence.

\subsection{Minimal Overhead of Multi-Resolution Convolution}
Following the state-space block, our model applies a multi-resolution convolution step, which similarly exhibits a computational complexity that scales linearly or near-linearly with the chunk size $c$. Each parallel convolution within this module utilizes a small kernel (e.g., of size 3) with varying dilation factors. Even if we employ multiple parallel convolutional branches, each convolution operation runs in $\mathcal{O}(c \cdot d)$, and any additional overhead associated with the summation or gating of their outputs remains minor compared to the cost of computing a full attention map. Consequently, these convolutions do not reintroduce any significant quadratic terms into the overall complexity, and the memory usage remains bounded by the per-chunk dimension $c \times d$.

\subsection{Efficient Information Retrieval from External Memory}
One potential concern might be the accumulation of an ever-growing store of chunk embeddings in our external memory module as we process more chunks. However, rather than storing all the token embeddings from all the processed chunks, we only store a single summary vector (for example, a pooled embedding) for each chunk, which drastically compresses the amount of stored data. When we need to access information from past chunks, we retrieve only a small number (e.g., the top-$k$) of relevant keys and their corresponding values based on a nearest neighbor search. Implementations of approximate nearest neighbor indexing, such as those using the Facebook AI Similarity Search (FAISS) library (Johnson et al., 2017), typically exhibit a search complexity of $\mathcal{O}(\log E)$ or $\mathcal{O}(\mathrm{polylog}\, E)$, where $E$ is the number of entries stored in the memory. Even for tens or hundreds of thousands of chunks, this is significantly more scalable than computing a pairwise attention matrix. The memory overhead associated with storing the keys and values in the external database is separate from the model's forward-pass activation footprint, and the retrieval process does not require any $n \times n$ operations. Thus, the cost of retrieval grows sub-linearly with the number of chunks, rather than quadratically with the number of tokens.

\subsection{Negligible Cost of the Global Recurrent Supervisor}
To facilitate information flow across segments in a more direct manner than retrieval alone, we incorporate a global recurrent cell (for example, a GRUCell). This cell updates a single (or a small set of) hidden state vector(s) after processing each chunk. Because the size of this hidden state is typically on the order of $d_{\mathrm{rnn}}$ (which is much smaller than the chunk size $c$), the computational overhead associated with the recurrent cell is on the order of $\mathcal{O}(B \cdot d_{\mathrm{rnn}})$ per chunk for a batch size $B$. Consequently, the cost added by the recurrent state is negligible relative to the chunk-level convolution or state-space operations, and the memory overhead is simply $\mathcal{O}(B \times d_{\mathrm{rnn}})$. Crucially, this cost does not explode with the total sequence length $n$, as we only need to store one such state per chunk as we progress through the input. The additional overhead of maintaining the recurrent cell is significantly outweighed by the savings achieved by completely removing the attention mechanism.

\subsection{Comparison with Alternative Long-Context Methods}
Some efficient-attention approaches also claim sub-quadratic or even linear computational complexity. For instance, BigBird's local attention mechanism can scale in $\mathcal{O}(n)$ if the size of the local window is fixed, and Performer's kernel expansions can achieve linear attention. However, these solutions typically still conceptualize the interaction between tokens based on pairwise attention, using either sparse patterns or kernel-based approximations. Scaling these methods beyond even 100,000 tokens often reveals latent constant factors or practical overheads, such as the need for large hidden state buffers or complex management of sparse attention blocks. In contrast, our state-space and convolution modules inherently avoid token-to-token attention. This leads to simpler scaling properties in practice, as we never need to allocate any matrix whose size grows quadratically with $n$. Instead, each chunk is processed independently, with the bridging of information across chunks handled by the retrieval mechanism and a small recurrent network.

\subsection{Practical Resource Allocation and Efficiency}
In practical training and inference scenarios, we can further optimize the resource usage by pipelining or streaming the chunk operations. The model can process chunk $m$ while chunks $m-1$ and $m+1$ are inactive, effectively keeping only the activations of the current chunk, the global recurrent state, and any necessary retrieval embeddings in the GPU memory at any given time. At inference time, this approach allows us to process arbitrarily long sequences (up to the capacity of the external memory) without a significant spike in the GPU or TPU memory footprint. The absence of an $n \times n$ tensor proves particularly advantageous for tasks such as summarizing entire books, analyzing large codebases, or scanning multi-document datasets, where attention-based solutions would typically encounter memory limitations or require prohibitively long computation times.

\subsection{Summary of Complexity and Memory Gains}
In summary, while self-attention yields a computational complexity and memory usage of $\mathcal{O}(n^2)$, making it impractical for million-token contexts, our non-attention-based model fundamentally shifts this paradigm by combining:

\begin{itemize}
    \item Chunked computation with a complexity of $\mathcal{O}(c)$ or $\mathcal{O}(c \log c)$ per chunk using state-space blocks and multi-resolution CNNs.
    \item External memory with a retrieval cost that is sublinear or logarithmic in the number of stored chunk embeddings.
    \item A global recurrent update that operates in $\mathcal{O}(1)$ or $\mathcal{O}(B)$ time per chunk.
\end{itemize}

Therefore, across a sequence of $n$ tokens, the total computational cost of our model is approximately $\mathcal{O}(n \log c)$ or $\mathcal{O}(n)$, and the memory footprint is linear in the chunk size $c$ plus some overhead for local convolution and retrieval. This significant shift from $\mathcal{O}(n^2)$ to near-linear scaling represents the core benefit of completely discarding the attention mechanism in favor of a design based on state-space models, convolution, and retrieval.

\section{Experimental Results: Validating Performance and Scalability}

\subsection{Overview of Evaluation Strategy}
To comprehensively evaluate the performance and scalability of our proposed non-attention-based architecture, we conducted a series of experiments across a range of language modeling and long-context tasks. Our primary objective was to demonstrate two key aspects of our model:

\begin{itemize}
    \item Competitive Performance on Standard Benchmarks: We aimed to show that our model can achieve performance levels comparable to strong baseline models on traditional language modeling metrics, such as perplexity (PPL) on well-established benchmark datasets.
    \item Robust Handling of Ultra-Long Sequences: We sought to demonstrate the model's ability to effectively process and reason over extremely long sequences (on the order of 100,000 to 1 million tokens), a regime where conventional self-attention Transformer models either fail to scale due to computational and memory limitations or require prohibitively large computational resources.
\end{itemize}

To this end, we report results on standard mid-sized benchmark datasets, including WikiText-103 and Enwik8, as well as on synthetic and real-world tasks specifically designed to test the model's capacity for handling extended context windows. All experiments were conducted on GPU clusters with sufficient memory to allow for direct comparisons between different methods, ensuring a fair environment for measuring runtime and memory usage.

\subsection{Datasets and Benchmarks Used in Our Evaluation}
In our core evaluations, we utilized the following datasets:

\begin{itemize}
    \item WikiText-103: A widely used benchmark dataset for language modeling, consisting of over 100 million tokens extracted from curated Wikipedia articles. We employed the standard training and test splits and measured the validation and test perplexity achieved by our model.
    \item Enwik8: A byte-level benchmark dataset from the large text compression community, comprising 100 million characters of English Wikipedia text. Models are typically evaluated on this dataset using the bits per character (bpc) metric, which rewards fine-grained modeling of sub-token structures.
    \item Long Synthetic Text (LST) Corpus: We constructed a synthetic dataset containing artificially generated sequences of varying lengths, up to 500,000 tokens, specifically designed to stress-test each model's ability to maintain coherence and track dependencies over extremely long contexts. For instance, each text in this corpus might contain repeated patterns or cross-segment references that require the model to remember and relate information across tens of thousands of tokens.
    \item Book-Level Summaries: We also evaluated our model on a smaller, curated set of publicly available books (sourced from Project Gutenberg) with lengths ranging from 50,000 to 200,000 tokens each. We tested the model in a summarization or "predict the next chunk" style task, which reflects real-world use cases such as analyzing entire novels or lengthy legal documents in a single pass.
\end{itemize}

By evaluating our model across these four diverse datasets, we aimed to assess not only its raw language modeling performance on standard benchmarks but also its capacity to retain and effectively leverage contextual information across very large textual spans.

\subsection{Baseline Models and Compared Methods}
To provide a meaningful context for our results, we compared the performance of our proposed architecture against several well-established baseline models:

\begin{itemize}
    \item Vanilla Transformer: We included a standard Transformer model (similar to the GPT-2 architecture) with a maximum context window of 1024 or 2048 tokens, scaling up the model size where feasible. We observed that running these models on sequences exceeding these lengths often became computationally infeasible or required the use of large model-parallel clusters at context lengths of 100,000 tokens or more.
    \item Efficient Transformers: We compared our model against prominent efficient Transformer variants, including BigBird and Longformer, which employ local and global attention mechanisms, as well as Performer, which utilizes kernel-based linearized attention. Each of these baseline models was configured with hyperparameters recommended for handling extended context, typically up to 32,000 or 64,000 tokens in practice.
    \item RWKV: We also included RWKV, a recently proposed recurrent-based model that claims linear scaling with respect to context length. While RWKV can theoretically handle very long sequences, we found that careful hyperparameter tuning was essential to achieve competitive performance on standard benchmark datasets.
    \item S4-Only Variant: To isolate the impact of the different components of our full pipeline, we also tested a pure "S4 chunked" model that did not include retrieval augmentation or the global recurrent supervisor. This allowed us to demonstrate the incremental value contributed by each of these components.
\end{itemize}

\subsection{Implementation Details and Hyperparameter Settings}
For our proposed model, we segmented the input sequences into chunks of size $c = 1024$ for tasks that did not require ultra-long contexts. For larger-scale trials involving very long sequences, we used chunk sizes of $c = 2048$ or $c = 4096$. Each state-space block in our model utilized kernel sizes of 16 or 32 for the simplified S4-like convolution approach, along with an embedding dimension ranging from 256 to 512, depending on the specific dataset. The multi-resolution convolution module employed 2 to 3 parallel convolutional branches with different dilation factors (for example, [1, 2, 4]). For the retrieval mechanism, we stored a single pooled chunk embedding of dimension 256 for each chunk in an external memory, using a small FAISS index for efficient approximate nearest-neighbor search. We typically retrieved the top-1 or top-2 most similar vectors and fused them back into the current chunk representation. The global recurrent supervisor was implemented as a single-layer GRUCell or LSTMCell with a hidden size of 512.

The optimization process followed a standard language modeling protocol, using the AdamW optimizer with a linear warm-up of the learning rate followed by a decay schedule. We used mini-batches consisting of 8 to 16 chunks, depending on the available GPU memory. For evaluation, each model generated or validated over the entire sequence in a chunked, autoregressive fashion.

\subsection{Performance on WikiText-103 and Enwik8 Benchmarks}
We first present the results of our model on classical language modeling benchmarks to establish that it can achieve competitive accuracy, independent of its scalability advantages for very long contexts. Table 1 summarizes the test perplexity (PPL) achieved on the WikiText-103 dataset and the bits per character (bpc) on the Enwik8 dataset.

\begin{itemize}
    \item WikiText-103: Our proposed model achieved a test perplexity of approximately 18.7 on the WikiText-103 dataset. This performance slightly outperforms efficient Transformer models like BigBird (which achieved a perplexity of 19.2 in our experiments) and is on par with or better than a kernel-based Performer model under similar parameter budgets (approximately 19.0 to 19.5). A vanilla Transformer model of comparable size yielded a perplexity of around 20.5, which we attribute to the model's limited ability to fully exploit the context at standard window sizes of 1024 to 2048 tokens.
    \item Enwik8: On the Enwik8 byte-level benchmark, our model achieved a performance of 1.04 bits per character (bpc). This result is comparable to strong Transformer baselines, which typically achieve bpc scores in the range of 1.05 to 1.06. The combination of the multi-resolution convolution and the state-space block in our architecture appeared to effectively model the local byte patterns, while the chunk-level recurrence mitigated the need for attention at the character-to-character level.
\end{itemize}

These results on standard benchmark datasets confirm that the absence of a self-attention mechanism in our proposed pipeline does not noticeably hinder its performance on traditional language modeling tasks. On the contrary, the synergistic interaction between state-space mixing, retrieval augmentation, and global recurrence often led to slight improvements in performance compared to models relying on purely local or approximate attention mechanisms.

\subsection{Scaling Performance to Ultra-Long Contexts}
To truly evaluate the scalability of our approach, we conducted experiments on the Long Synthetic Text (LST) corpus and the Book-Level Summaries task, where the input sequences could extend from 50,000 up to 500,000 or even 1 million tokens. In many of these scenarios, standard Transformer models either failed to run due to memory limitations or required the use of specialized hardware or a reduction in batch size to effectively 1, which drastically slowed down training and often degraded performance.

\begin{itemize}
    \item Long Synthetic Text (LST): Our proposed model demonstrated linear or near-linear scaling with respect to the sequence length, successfully training on sequences ranging from 100,000 to 200,000 tokens in a chunked manner. In a repeated pattern detection experiment within this corpus, we observed that the global recurrent cell and the retrieval-based external memory enabled the model to accurately recall a pattern that had been introduced up to 100,000 tokens earlier in the sequence. Comparisons with BigBird and Longformer at a context length of 64,000 tokens showed that their local attention windows occasionally missed certain references that were spread across multiple chunks, unless the window size was very carefully tuned.
    \item Book-Level Summaries: When tasked with predicting or summarizing entire public domain novels with lengths ranging from 50,000 to 200,000 tokens, our proposed approach consistently outperformed simpler chunked baseline models that lacked either the retrieval mechanism or the global RNN. Notably, our model did not exhibit exponential growth in memory usage, as each chunk was processed in near $\mathcal{O}(c)$ time, and the global bridging of information occurred either through the recurrent network or the key-value memory. In contrast, attempts to run a naive Transformer model on the entire novel at once proved infeasible beyond a context length of roughly 8,000 tokens on a single GPU.
\end{itemize}

\subsection{Runtime and Memory Analysis in Practical Settings}
In addition to evaluating the accuracy metrics, we also tracked the wall-clock time and peak GPU memory usage of our model across different context lengths. For instance, when using a chunk size of 2048 and processing a total sequence length of 32,000 tokens, our model typically ran 1.5 to 2 times faster than an efficient Transformer variant like Longformer. As we increased the sequence length to 100,000 tokens or more, this disparity in runtime became even more pronounced, with naive self-attention baselines either hitting memory limits or taking an extremely long time to compute the partial attention blocks. Our retrieval memory mechanism used approximately 1 to 2 GB of storage for the chunk embeddings, but this memory cost was separate from the direct GPU memory footprint during the forward and backward passes. Overall, the near-linear chunk-based approach of our model never required the allocation of a buffer whose size scaled with $(100,000)^2$, thus avoiding the catastrophic memory explosion observed in classical Transformer layers when dealing with such long sequences.

\subsection{Ablation Studies: Understanding the Contribution of Each Component}
To better understand the contribution of each component in our proposed architecture to its overall performance on long-context tasks, we conducted a series of ablation experiments:

\begin{itemize}
    \item No State-Space (Conv-Only): When we replaced the S4-like block with plain convolutional layers at the chunk level, we observed a noticeable worsening of both perplexity and text coherence, particularly on the extremely long sequences in our synthetic dataset. This suggests that the S4-like approach was crucial for capturing broad, intra-chunk patterns effectively.
    \item No Retrieval: Removing the external memory store from our architecture led to a significant decline in performance on tasks that required the model to recall information from older parts of the sequence, such as cross-chapter references in books or repeated patterns in the synthetic text corpus.
    \item No Global RNN: Eliminating the global recurrent supervisor resulted in chunk boundaries becoming a more significant barrier to information flow. The model had to rely solely on the retrieval mechanism for bridging information across chunks, which still provided some benefit but proved less effective when the nearest neighbors in the memory were noisy or too far removed in the sequence.
    \item Full Pipeline: The complete architecture, incorporating the S4-like block, multi-resolution CNN, retrieval augmentation, and the global RNN, consistently provided the best trade-off between perplexity and stable learning across extended context lengths in our experiments.
\end{itemize}

\subsection{Summary of Findings}
Overall, these experimental results demonstrate that our non-attention LLM can achieve competitive language modeling quality on standard benchmarks, while scaling smoothly to contexts that are prohibitively large for typical transformers. The chunk based processing flow, state space mixing, multi-scale convolution, retrieval augmentation, and lightweight recurrence collectively enable the model to handle 100k+ tokens with stable training, moderate GPU memory usage, and high-quality text generation. This suggests a practical route to genuine million token contexts—where most alternative approaches either revert to approximate or partial attention or require vast compute clusters just to store intermediate states.

\begin{table}[h!]
\centering
\begin{tabular}{|p{4cm}|p{2cm}|p{2cm}|p{2cm}|p{2cm}|} 
\hline
\bf{Model} & \bf{Params} & \bf{Context} & \bf{WikiText-103  (PPL)} & \bf{Enwik8 (bpc)} \\ \hline
GPT-2 Small  (baseline) & 124M & 1024 & 20.5 & 1.10 \\ \hline
BigBird  (Eff. Trans.)1 & 110M & 4096 & 19.2 & 1.05 \\ \hline
Proposed (SSM+Conv  & 120M & 32k (chunk)  & 18.7 & 1.04\\ \hline
RWKV & 90M & unlimited  & 19.5 & 1.06  \\ \hline
S4-only  (no memory) & 100M  & 8k & 19.0 & 1.07 \\ \hline
\end{tabular}
\caption{ Perplexity Comparison where Params indicates approximate parameter counts for each model, WikiText-103 (PPL): the lower the perplexity, the better and Enwik8 (bpc): bits per character, where smaller is better.}
\label{tab:my_5x6_table}
\end{table}

The proposed method achieves perplexities competitive with or slightly better than specialized efficient transformers, especially as context grows beyond standard lengths (such as  8k+ tokens).

The results presented in Table \ref{tab:my_5x6_table} demonstrate that our proposed method achieves perplexity scores on WikiText-103 that are competitive with or slightly better than specialized efficient Transformer models, particularly as the effective context window grows beyond standard lengths (such as 8,000+ tokens). Similarly, the bits per character achieved on Enwik8 are comparable to strong Transformer baselines, further validating the effectiveness of our non-attentional approach on standard language modeling tasks.

\section{Conclusion and Directions for Future Research}

In this paper, we have presented a novel non-attention-based architecture for large language models that is capable of processing contexts on the order of millions of tokens without incurring the prohibitive computational costs associated with attention-based Transformer networks. Our design fundamentally eschews the conventional $\mathbf{QK}^\top\mathbf{V}$ self-attention mechanism, which has long been recognized as a primary source of $\mathcal{O}(n^2)$ complexity, and instead integrates four complementary modules into a cohesive processing pipeline. First, a state-space block, inspired by the S4 model, efficiently models both short-range and mid-range dependencies within each chunk of the input sequence in near-linear time. Second, a multi-resolution convolution layer captures local structural information through the application of dilated kernels at multiple scales. Third, a global recurrent supervisor, such as a GRUCell, maintains continuity and coherence across the processed chunks by passing along a compact hidden state. Finally, a retrieval-augmented external memory provides a mechanism for storing and retrieving embeddings of past segments of the input without the memory usage scaling quadratically with the total sequence length.

Collectively, these components effectively address the challenge of representing and reasoning over extremely long sequences in a manner that is significantly more efficient than purely attention-based methods. Our experimental results indicate that, even on standard language modeling benchmarks like WikiText-103 and Enwik8, the proposed model achieves competitive or even superior perplexity and bits per character metrics compared to Transformer models that employ sparse or approximate attention mechanisms. When applied to ultra-long contexts, such as synthetic text corpora, entire book-length documents, or extensive codebases, our chunk-based state-space design remains computationally tractable, whereas naive self-attention approaches quickly exhaust available GPU memory or necessitate the use of highly specialized (and costly) hardware solutions. By neatly separating the local processing of information (via SSMs and convolution) from the high-level integration of information across chunks (via a small recurrent cell and retrieval), our model avoids the need to construct massive $n \times n$ attention matrices, leading to more stable and efficient scaling in both training and inference.

Despite these promising outcomes, several avenues remain for future research and development. One immediate direction is to implement a fully optimized version of the S4 layer within our architecture, leveraging the latest kernel generation methods and GPU-accelerated libraries. This could potentially further reduce the computational overhead associated with processing each chunk and improve the model's ability to capture long-range dependencies within those chunks. Another interesting possibility is to explore the use of adaptive chunking strategies: instead of relying on a fixed chunk size $c$, the model could dynamically segment the input based on semantic or syntactic cues, potentially leading to improvements in both coherence and memory usage. We also envision the development of more sophisticated retrieval mechanisms, wherein the external memory is learned or fine-tuned to store hierarchical representations of the input, or where the model can integrate the retrieved signals more selectively. In particular, the introduction of a learned attention-like gating mechanism (distinct from token-level attention) could refine how the retrieved embeddings are fused with the local chunk representation. Finally, evaluating the performance of our model on a broader range of tasks beyond language modeling, such as multi-document summarization, legal contract analysis, or large-scale code refactoring, could serve as real-world testbeds for further verifying the system's capacity to handle massive context windows in a robust and interpretable manner.

In summary, by relinquishing the reliance on pairwise token interactions, our proposed approach demonstrates a compelling conceptual alternative to Transformer networks that prioritizes scalability, modularity, and interpretability over a traditionally monolithic attention mechanism. As the demand continues to grow for models capable of ingesting and reasoning over increasingly large sequences of information, we believe that this non-attention-based blueprint opens up a valuable new research horizon, one where explicit state-space modeling, local convolution, memory retrieval, and global recurrence can form a powerful and cohesive system for next-generation language understanding and generation. The code used to train and evaluate our models is available at \url{https://github.com/andrew-jeremy/nonAttentionLLM}.

\vskip 0.2in
\bibliographystyle{unsrt}
\bibliography{references}

\end{document}